\documentclass{article}
\usepackage{arxiv}
\usepackage{graphicx} 

\usepackage{amsmath,amsfonts,bm}

\def\eqref#1{equation~\ref{#1}}

\def\1{\bm{1}}

\DeclareMathAlphabet{\mathsfit}{\encodingdefault}{\sfdefault}{m}{sl}
\SetMathAlphabet{\mathsfit}{bold}{\encodingdefault}{\sfdefault}{bx}{n}

\usepackage{hyperref}
\usepackage{url}
\usepackage{subcaption} 
\usepackage{amsmath}
\usepackage{amsfonts}
\usepackage{eso-pic} 
\RequirePackage{fancyhdr}
\RequirePackage{natbib}

\usepackage{booktabs}
\usepackage{tabularx}

\title{Learning Compact Representations of LLM Abilities via  \\ Item Response Theory}
\author{
  \textbf{Jianhao Chen}$^{1~2~\ast}$ \quad 
  \textbf{Chenxu Wang}$^{2~3}$  \quad
  \textbf{Gengrui Zhang}$^{4}$ \quad
  \textbf{Peng Ye}$^{2}$ \\
  \textbf{Lei Bai}$^{2}$ \quad
  \textbf{Wei Hu}$^1$ \quad
  \textbf{Yuzhong Qu}$^{1~\dag}$ \quad
  \textbf{Shuyue Hu}$^{2~\dag}$ \quad \\
  $^1$ \text{State Key Laboratory for Novel Software Technology, Nanjing University} \\
  $^2$ \text{Shanghai Artificial Intelligence Laboratory}  \\
  $^3$ \text{Fudan University} \\
  $^4$ \text{Department of Psychology, University of Southern California} \\
  \texttt{jhchen.nju@gmail.com} \quad 
  \texttt{\{whu,yzqu\}@nju.edu.cn} \quad
  \texttt{gengruiz@usc.edu} \\
  \texttt{\{wangchenxu,yepeng,bailei,hushuyue\}@pjlab.org.cn} \\
}

\begin{document}
\renewcommand{\thefootnote}{}
\footnotetext{\dag~Corresponding author.}
\footnotetext{$\ast$~Work done during the author's intership at Shanghai Arificial Intelligence Laboratory.}
\footnotetext{$^1$~For example, NeurIPS 2025’s dataset and benchmark track received more than 1,900 submissions, and in 2024, more than 1,200 submissions; a large proportion of accepted submissions are LLM benchmarks.}
\maketitle

\begin{abstract}
Recent years have witnessed a surge in the number of large language models (LLMs), yet efficiently managing and utilizing these vast resources remains a significant challenge. In this work, we explore how to learn compact representations of LLM abilities that can facilitate downstream tasks, such as model routing and benchmark prediction. We frame this problem as estimating the probability that a given model will correctly answer a specific query. Inspired by the item response theory (IRT) in psychometrics, we model this probability as a function of three key factors:  (i) the model’s multi-skill ability embedding $\theta$, (ii) the query’s discrimination vector $\alpha$ that separates models of differing skills, and (iii) the query’s difficulty scalar $\beta$. To learn these parameters jointly, we introduce a Mixture-of-Experts (MoE) network that couples model- and query-level embeddings. Extensive experiments demonstrate that our approach leads to state-of-the-art performance in both model routing and benchmark accuracy prediction. Moreover, analysis validates that the learned parameters encode meaningful, interpretable information about model capabilities and query characteristics. Code and data are available at \url{https://github.com/JianhaoChen-nju/IrtNet}.
\end{abstract}

\section{Introduction}
Recent years have seen an explosion of large language models (LLMs), spanning from massive, general-purpose systems to small, task-specialized ones. As of August 2025, Hugging Face lists over 97,000 text-generation models with at least 6B parameters. This proliferation, however, introduces a significant challenge: how to \emph{efficiently} manage and utilize such a vast, rapidly expanding ecosystem. 

A crucial step toward addressing this challenge is to construct \emph{compact} representations of  models' abilities~\citep{embedllm}. By encoding each model's strengths and weaknesses, such representations can facilitate a range of downstream applications. For example, in \emph{model routing}, they allow for assessing a model’s suitability for a given query, so that queries can be assigned to the most appropriate model within a candidate pool~\citep{shnitzer2023large,zooter,jitkrittum2025universal}. This not only helps balance performance and cost~\citep{ong2024routellm,feng2024graphrouter,wang2025mixllm,frick2025prompt}, but also empowers ensembles of smaller models to effectively compete with large proprietary systems~\citep{zhang2025avengers,pan2025route}. Another use case is \emph{benchmark prediction}~\citep{polo2024tinybenchmarks,zhang2025benchmark}. Traditionally, benchmarks are designed to compare the abilities of different models on specific tasks. However, the recent surge in their number$^1$ makes it impractical for exhaustive evaluation. 
Compact representations of model abilities offer an alternative, enabling efficient, scalable LLM evaluation. 

\begin{figure}[t!]
    \centering
    \includegraphics[width=\columnwidth]{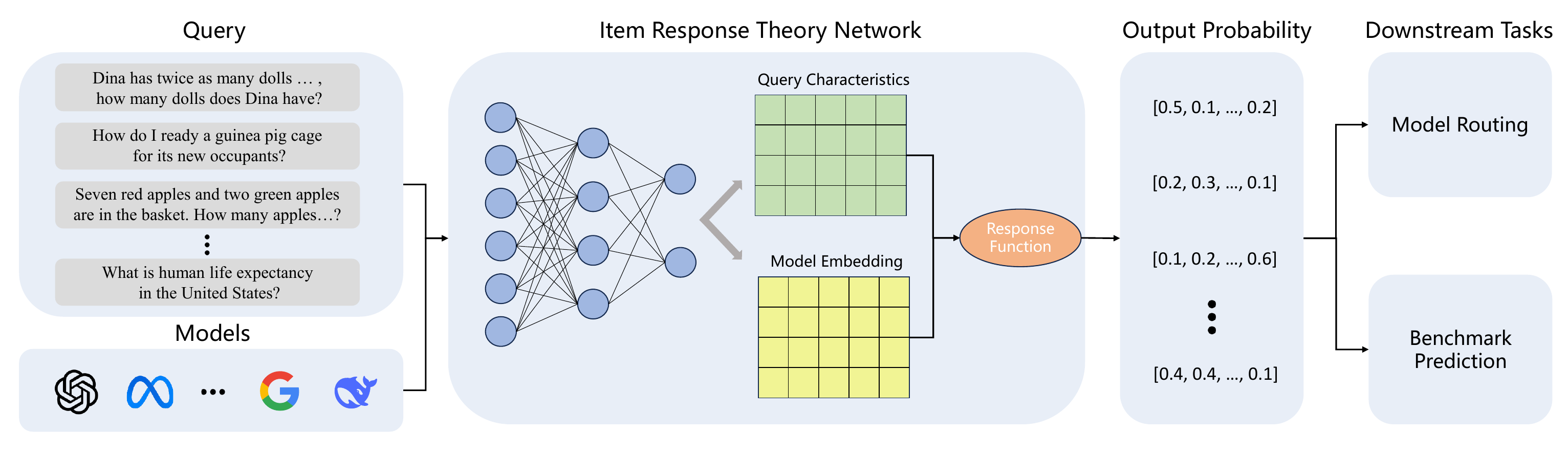}
    \caption{Overview of the IrtNet framework for learning LLM representations. IrtNet learns model embeddings based on models' past query answering performance and outputs probabilities that  models answer correctly. The output probability can be directly applied to downstream tasks containing model routing and benchmark prediction.}
    \label{fig:framework}
    \vspace{-10pt}
\end{figure}

In this work, we propose a novel approach to construct compact representations of model abilities as shown in Figure~\ref{fig:framework}, drawing inspiration from the item response theory (IRT). IRT is a well-established statistical framework used in education and psychology to measure latent abilities through standardized tests. It models the interactions between individuals' performance, their latent abilities, and query characteristics, such as difficulty and discrimination (as in the 2-parameter IRT). By treating queries as tests and LLMs as test respondents, we analogously model the likelihood of an LLM $m$ correctly answering a query $q$ as a mathematical function of $m$'s latent abilities $\theta_m$, query $q$'s difficulty $\beta_q$, and $q$'s discrimination power $\alpha_q$ between models. 
To estimate the model's compact representation of abilities ($\theta_m$) and the query characteristics ($\beta_q$ and $\alpha_q$), we present a neural network \emph{IrtNet}. This architecture enables end-to-end training, allowing for the simultaneous optimization of model latent abilities and query characteristics to align with the ground truth of whether a given model correctly answers a given query.
 
Comprehensive experiments show that our approach not only achieves state-of-the-art performance on downstream tasks but also produces interpretable learned representations. Specifically, in model routing, our method achieves an average accuracy of 67.4\%, significantly outperforming strong baselines like RouterDC~\citep{routerDC} (54.9\%), MODEL-SAT~\citep{modelSAT} (56.7\%), Avengers-Pro~\citep{zhang2025avengers} (62.1\%), EmbedLLM~\citep{embedllm} (60.2\%).  In benchmark prediction, our method demonstrates remarkable data efficiency. It reaches 69.9\% accuracy using less than 4\% of the training data, matching the state-of-the-art (EmbedLLM) performance achieved with the full training set. Moreover, our analysis validates that the learned parameters encode meaningful information: models' compact representations form clear clusters in the latent space based on model family (e.g., Llama, Qwen series) and specialization (e.g., models trained for coding and mathematics); the learned query difficulty parameter $\beta_q$ exhibits a near-perfect negative correlation with true benchmark scores, with a Pearson correlation coefficient of -0.9721. These findings indicate that IrtNet successfully captures the intrinsic model abilities and query characteristics, providing a powerful new tool for evaluation, selection, and management of the vast, rapidly expanding LLM ecosystem.


\section{Preliminary}
We denote the compact representation of an LLM $m$ by a $d$-dimensional embedding vector $\theta_m \in \mathbb{R}^d $, which encodes the model's strengths and weaknesses. 
Let $M=\{m_1,m_2,…,m_n\}$ be a set of $n$ distinct LLMs, and let $Q=\{q_1,q_2,…,q_k\}$ be a set of $k$ distinct queries. For any given model-query pair $(m,q)$, where $m\in M$ and $q\in Q$, the model's answer to the query can be represented by a binary outcome $y\in \{0,1\}$, where $y=1$ signifies a correct answer and $y=0$ signifies an incorrect one.

Our objective is to learn a probability mass function $f_{\theta}(m,q)=\text{Pr}(y=1|m,q)$, parameterized by $\theta=(\theta_1, \ldots, \theta_n)$, such that for any given model-query pair, the function $f_{\theta}$ can accurately predict whether the model can answer the query correctly or not. 
Once learned, the function $f_{\theta}$ can facilitate several important downstream tasks:



\textbf{Model Routing.~} 
Given a set of candidate models, model routing aims to assign each query to the most suitable model, avoiding the need for every model to answer it exhaustively~\cite{shnitzer2023large}. In this way, the ensemble of models effectively leverages collective intelligence, allowing the group to solve a broader range of tasks that any individual model alone cannot accomplish.
Formally, let $\mathcal{M} \subseteq M$ be the set of candidate models; based on the function $f_\theta$, for any query $q$, model routing can be achieved by selecting the model $m^\ast$ that yields the highest probability of generating a correct answer, i.e.,
    \begin{align}
    m^* = \underset{m \in \mathcal{M}}{\arg\max} f_\theta(m, q).
    \end{align}


\textbf{Benchmark Prediction.~} As exhaustive LLM evaluation is compute-intensive and time-consuming, benchmark prediction seeks to estimate the overall performance of an LLM from only a small subset of evaluation data~\citep{vivek2023anchor}. Formally, let $ \mathfrak{Q}$ be the set of queries that are not included when learning the function $f_\theta$, i.e., $\mathfrak{Q} \cap Q = \emptyset$, where typically $|\mathfrak{Q}| \gg |Q|$; for a given model $m$, benchmark prediction generates a predicted accuracy $\hat{S}$ for the set $\mathfrak{Q}$ by feeding each query from this set into the function $f_\theta$, which outputs the predicted probability of generating correct answers to these queries, i.e.,
    \begin{align}
    \hat{S} = \frac{1}{|\mathfrak{Q}|} \sum_{q \in \mathfrak{Q}} f_\theta(m, q).
    \end{align}
In practice, as $f_\theta$ is a neural predictor, a single pass through all queries in the set $\mathfrak{Q}$ can yield the predicted accuracy for all models in the set $M$. This avoids the need to run multiple passes (one pass for each model) as in traditional benchmarking LLMs, thereby allowing for efficient, scalable LLM evaluation.


Overall, establishing compact representations of LLMs provides a unified framework for understanding a model's strengths and weaknesses, supporting a range of downstream applications that have attracted significant recent interest. 


\section{Methodology}
\subsection{Modeling LLM Abilities via Item Response Theory}
The item response theory (IRT) is a well-established statistical framework to measure the latent abilities of respondents through testings~\citep{Cai2016item}. 
It is widely applied in education and psychology, and underpins the scoring scales of high-stakes exams such as the GRE and GMAT.

The IRT is based on the idea that the probability of a correct response to an item (or a question) is a mathematical function of a person's latent traits, indicating the person's ability, as well as  the item parameters.
Here, we focus on the two-parameter IRT model, which considers an item's difficulty and discrimination (its ability to differentiate between individuals)~\citep{reise2009item}.

Understanding LLM abilities through their performance on queries is analogous to assessing a person's abilities based on their performance on standardized tests. Thus, inspired by the IRT, we treat each LLM $m$ as a respondent with a latent trait, and each query $q$ as an item with two parameters that characterize its difficulty and discriminative power.

Formally,  let $\alpha_q \in \mathbb{R}^d$ denote  the query $q$'s $d$-dimensional discrimination parameter, and $\beta_q \in \mathbb{R}$ denote $q$'s difficulty parameter; we assume that the function $f_\theta(m,q)$, which predicts whether the model $m$ can correctly answer the query $q$, takes the following form of an item response function:
\begin{equation}
\label{eq:P}
f_\theta (m, q) = \sigma\!\left(\alpha_q^\top \theta_m - \beta_q \right)=\frac{1}{1+e^{-(\alpha_q^\top \theta_m-\beta_q)}},
\end{equation}
where $\theta_m \in \mathbb{R}^d$ is the model $m$'s compact representation (defined in Section 2), indicating its latent abilities, and $\sigma(\cdot)$ is the logistic link function. 
The discrimination parameter $\alpha_q$ suggests how important each latent ability dimension for the query is, where a higher value indicates that proficiency in that dimension is more critical for a correct answer.
The dot product $\alpha_q^\top \theta_m$ computes how well the model’s abilities align with the discriminative power of the query. Essentially, it measures the model’s fit to the query in terms of both the model’s ability in relevant latent dimensions (encoded by $\theta_m$) and the importance of those dimensions to the query (encoded by $\alpha_q$).

\subsection{IrtNet: Learning Compact Representations}
We propose to jointly learn models' compact representations $\theta=(\theta_1, \ldots, \theta_n)$, queries' discrimination parameters $\alpha= (\alpha_1, \ldots, \alpha_k)$, and difficulty parameters $\beta= (\beta_1, \ldots, \beta_k)$ through a neural network \emph{IrtNet} in an end-to-end manner.

\begin{figure}[t!]
    \centering
    \includegraphics[width=\columnwidth]{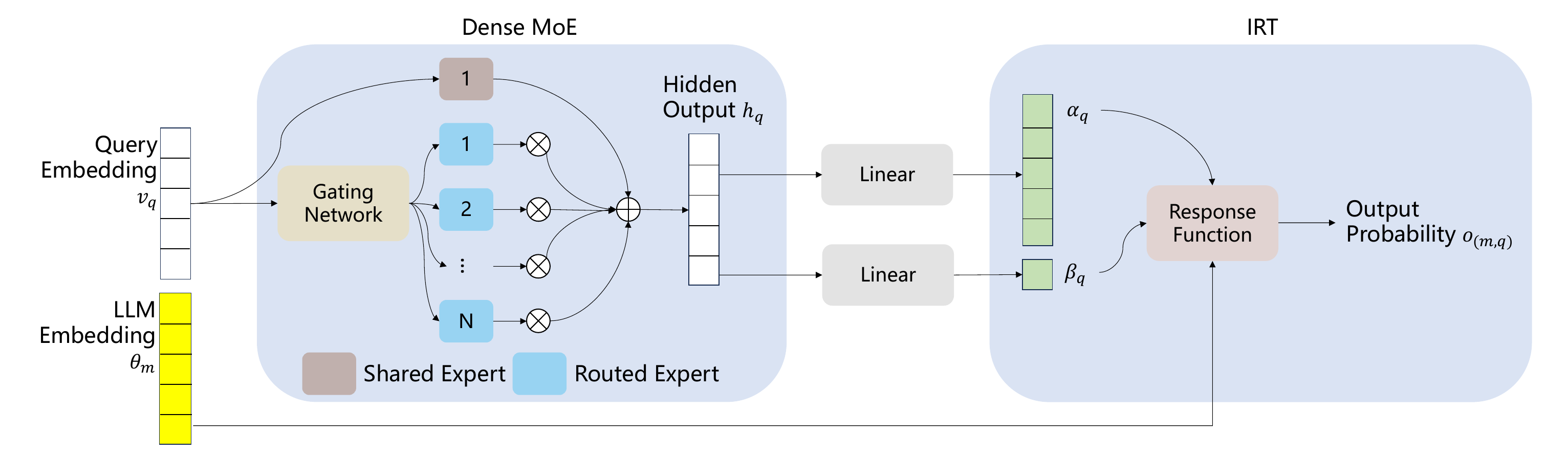}
    \caption{The architecture of IrtNet. A query embedding is processed through a dense MoE layer and subsequent linear layers to generate the query's discrimination $\alpha_q$ and difficulty $\beta_q$ parameters. These parameters are then combined with an LLM embedding $\theta_m$ via the response function to compute the final output probability.}
    \label{fig:irtnet}
\end{figure}

We illustrate the architecture of IrtNet in Figure~\ref{fig:irtnet}. First, each query $q$ is converted into a semantic embedding $v_q$ using a pre-trained embedding model. Next, $v_q$ is fed into the MoE layer of the IrtNet. We use a dense MoE layer which always activates all $N$ experts, with an auxiliary-loss-free load balancing strategy~\citep{liu2024deepseek} to promote a diverse weight distribution. This design aims to capture diverse, multi-faceted understanding of the query, improving prediction accuracy while avoiding the training instability problem often encountered in sparse MoE layers.
The query's hidden output $h_q$ is obtained by
\begin{align}
    h_q=\text{MoE}(v_q)= \text{SharedExpert}(v_q)+\sum_{i=1}^{N}w_i\cdot \text{RoutedExpert}_i(v_q).
\end{align}
Next, we employ two independent linear layers to obtain the discrimination parameter $\alpha$ and difficulty parameter $\beta$:
\begin{align}
   \alpha_q=\text{Linear}(h_q, d), \quad 
   \beta_q=\text{Linear}(h_q, 1).
\end{align}
where $d$ is the dimension of $\theta_m$.
Finally, the IrtNet combines the model embedding $\theta_m$ with the query characteristic parameters $\alpha_q$ and $\beta_q$ to compute the output $o_{(m,q)}$ based on the response function shown in Equation~\ref{eq:P}.

To learn these parameters, we define the objective as minimizing the discrepancy between the predicted probabilities $o_{(m,q)}$ and the ground-truth labels $y$ across the entire training dataset. Specifically, for a training set $\mathcal{D}$ consisting of samples $(m, q, y)$, the overall loss $\mathcal{L}$ is formulated as the sum of the binary cross-entropy losses for all samples:
\begin{align}
    \mathcal{L} = - \sum_{(m,q,y) \in \mathcal{D}} [y \log(o_{(m,q)}) + (1-y) \log(1 - o_{(m,q)})].
\end{align} 

\section{Experiments}
\subsection{Experimental Setup}
\paragraph{Datasets}
We use the same data as EmbedLLM~\citep{embedllm}. We applied a majority vote to consolidate multiple answers from a model to the same query. This step ensures a unique ground truth for each model-query pair, which is especially critical for the test set. The datasets contain 35,673 queries from 10 public benchmarks, including ASDiv~\citep{asdiv}, GPQA~\citep{gpqa}, GSM8K~\citep{gsm8k}, MathQA~\citep{mathqa}, LogiQA~\citep{logiqa}, MedMCQA~\citep{medmcqa}, MMLU~\citep{MMLU}, SocialIQA~\citep{SocialIQA}, PIQA~\citep{piqa}, and TruthfulQA~\citep{truthfulQA}. The correctness of answers from 112 open-source language models to those queries  was evaluated. The queries were converted into 768-dimensional embeddings using the all-mpnet-base-v2~\citep{reimers2019sentence} sentence transformer. The queries were split into a training set of 29,673 queries, a validation set of 3,000 queries, and a test set of 3,000 queries.

\paragraph{Hyperparameters}
The dimension $d$ of the compact LLM representation $\theta_m$ is 232. The number of experts in the MoE layer is set to 40, which is 4 times the number of datasets.

\subsection{Model Routing}
In this section, we test IrtNet's ability to serve as an effective and intelligent router in a multi-model environment. The objective is to leverage the fine-grained predictions from our framework to select the best model from a diverse pool to handle a given query, thereby maximizing both accuracy and efficiency. For any given query $q$, we compute the predicted success probability $P(y=1|m,q)$ for all candidate models and route the query to the model with the highest probability.
\begin{table}[!h]
    \centering
    \small
    \caption{Model routing accuracy (\%) comparison. \textbf{Micro} denotes the micro-averaged accuracy, calculated on the total correct predictions across all datasets. \textbf{Macro}  denotes the macro-averaged accuracy, computed by first calculating the accuracy for each benchmark individually and then averaging these benchmark scores. We use \textbf{bold} to indicate the best results. }
    \begin{tabularx}{\textwidth}{lc*{12}{>{\centering\arraybackslash}X}}
        \toprule
        \bf{Method} & \bf{ASD} & \bf{GPQA} & \bf{GSM} & \bf{Math} & \bf{Logi}  & \bf{Med} & \bf{Mmlu}  & \bf{Social} & \bf{PIQA} & \bf{Truth} & \textbf{Micro} & \textbf{Macro} \\
        \midrule
        
        RouterDC & 62.1 & 21.2 & 70.5 & 40.5 & 41.2 & 54.5 & 70.9 & 30.9 & 80.6 & 50.0 & 54.9 & 52.2 \\
        EmbedLLM & 34.9 & 24.8 & 82.1 & 47.7 & 31.4  & 65.8 & 80.1 & \bf{37.0} & 86.6  & \bf{52.7}  & 60.2 & 54.3 \\
        MODEL-SAT & 6.57 & 22.6 & 80.4 & 43.0 & 47.0 & 62.4 & 80.8 & 29.6 & 83.6 & 35.1  & 56.7 & 49.1
        \\
        Avengers-Pro & 12.1 & \bf{29.4} & \bf{89.3} & 55.3 & \bf{49.0}  &  72.9 & 83.2 & 30.9 & 86.6 & 41.9 & 62.1 & 55.7 \\
        IrtNet (Ours) & \bf{66.7} & 28.6 & \bf{89.3} & \bf{57.8} & \bf{49.0}  & \bf{74.0} & \bf{86.2} & 34.0  & \bf{87.3} & 47.3 & \bf{67.4} & \bf{62.0} \\
        \bottomrule
    \end{tabularx}
    \vspace{-5pt}
    \label{tab:performance_comparison}
\end{table}

We compare our method with four advanced routing methods, RouterDC~\citep{routerDC}, EmbedLLM~\citep{embedllm}, MODEL-SAT~\citep{modelSAT} (with Qwen3-0.6B-Base~\citep{yang2025qwen3} as base model), and Avengers-Pro~\citep{zhang2025beyond}. As shown in Table~\ref{tab:performance_comparison}, our method achieves the best performance on both micro-average and macro-average metrics across 10 benchmarks. Specifically, IrtNet achieves a final micro-average accuracy of 67.4\%, significantly outperforming strong baselines like EmbedLLM (60.2\%) and Avengers-Pro (62.1\%), which showcases its immense potential for the model routing task.

\subsection{Benchmark Prediction}
We use two settings to test the benchmark prediction ability of IrtNet: in-distribution (ID) and out-of-distribution (OOD). In the ID setting, while the two sets $\mathfrak{Q}$ and $Q$ have no overlap, they may still be closely related. In the OOD case, performance on a dataset may be predicted based on the model’s performance on a different dataset.

\textbf{ID Correctness Prediction.~} 
In this setting, we evaluate IrtNet's core capability for correctness prediction on model-query pairs. The results, presented in Table~\ref{tab:correctness}, highlight our model's exceptional data efficiency and predictive power. 
\begin{table}[!h]
    \centering
    \caption{Correctness prediction accuracy (\%) on different training data sizes.}
    \label{tab:correctness}
    \begin{tabularx}{\textwidth}{lc*{7}{>{\centering\arraybackslash}X}}
        \toprule
        \multicolumn{1}{l}{\textbf{method}} & \multicolumn{7}{c}{\textbf{Dataset Size}} \\
        \cmidrule(lr){2-8}
         & \textbf{1K} & \textbf{5K} & \textbf{10K} & \textbf{15K} & \textbf{20K} & \textbf{25K} & \textbf{Full (29K)} \\
        \midrule
        KNN & 62.6 & 63.0 & 64.4 & 65.1 & 64.6 & 64.4 & 64.7 \\
        EmbedLLM & 60.8 & 64.2 & 66.5 & 67.9 & 69.1 & 69.9 & 70.6 \\
        IrtNet (ours) & \bf{69.9} & \bf{71.5} & \bf{71.7} & \bf{71.8} & \bf{72.0} & \bf{72.1} & \bf{72.2} \\
        \bottomrule
    \end{tabularx}
    \vspace{-5pt}
\end{table}

A key advantage of our framework is its ability to achieve strong performance with remarkably little training data. With just 1,000 queries—less than 4\% of the full training set—IrtNet achieves a prediction accuracy of 69.9\%. This performance significantly surpasses both the traditional KNN baseline (62.6\%) and the state-of-the-art EmbedLLM (60.8\%) trained on the same amount of data. Notably, our model's accuracy with only 1K queries already approaches the performance of EmbedLLM trained on the entire dataset (70.6\%). 

As the training set size increases, IrtNet consistently outperforms other baselines, reaching a final accuracy of 72.2\% on the full training set.  This demonstrates that our framework not only learns rapidly from limited samples but also scales effectively. The results strongly suggest that by modeling the interaction between model abilities and query characteristics, IrtNet captures their complex relationship more efficiently and accurately than existing methods. 

\textbf{OOD Benchmark Prediction.~} In this setting, we test IrtNet's ability to generalize its learned representations to make macroscopic performance predictions. The objective is to forecast an LLM's accuracy score on an entire benchmark it has never seen during training. This experiment uses a leave-one-out approach: we train IrtNet on all data except for one target benchmark and then use the trained model to predict the overall accuracy of all LLMs on that held-out benchmark. We exclude ASDiv and SocialIQA from this analysis, as the near-uniform scores across all models (approximately 0 and 0.3, respectively) suggest the data might represent noise rather than meaningful performance variation. This OOD setup serves as the ultimate test of whether our model representations can truly understand and generalize the abstract concept of LLM abilities.
\begin{table}[!h]
    \centering
    \small
    \caption{Root mean square error (RMSE) for OOD benchmark prediction. \textbf{Overall} denotes the total prediction error, calculated by treating all benchmarks as a single, unified test set.}
    \begin{tabularx}{\textwidth}{l*{9}{>{\centering\arraybackslash}X}}
        \toprule
        \bf{Method}  & \bf{GPQA} & \bf{GSM}  & \bf{Math} & \bf{Logi} & \bf{Med} & \bf{Mmlu} & \bf{PIQA} &  \bf{Truth} & \textbf{Overall}  \\
        \midrule
        EmbedLLM & 0.26  & 0.20  & 0.09 & 0.11 & 0.04 & 0.25 & 0.40 & \bf{0.11} & 0.21 \\
        IrtNet (Ours) & 0.26   & \bf{0.19}  & 0.09 & 0.11 & 0.04 & \bf{0.19} & \bf{0.36} & 0.12 & \bf{0.19} \\
        \bottomrule
    \end{tabularx}
    \label{tab:benchmarkPrediction}
    \vspace{-5pt}
\end{table}

The results in Table~\ref{tab:benchmarkPrediction} demonstrate IrtNet's robust benchmark accuracy prediction. Across the eight benchmarks, IrtNet achieves a lower root mean square error (RMSE) than EmbedLLM on three datasets and performs equally on four. This leads to an overall RMSE of 0.19, representing a nearly 10\% error reduction over EmbedLLM's 0.21.

\begin{figure}[htbp]
    \centering
    \begin{subfigure}[b]{0.32\columnwidth}
        \centering
    \includegraphics[width=\linewidth]{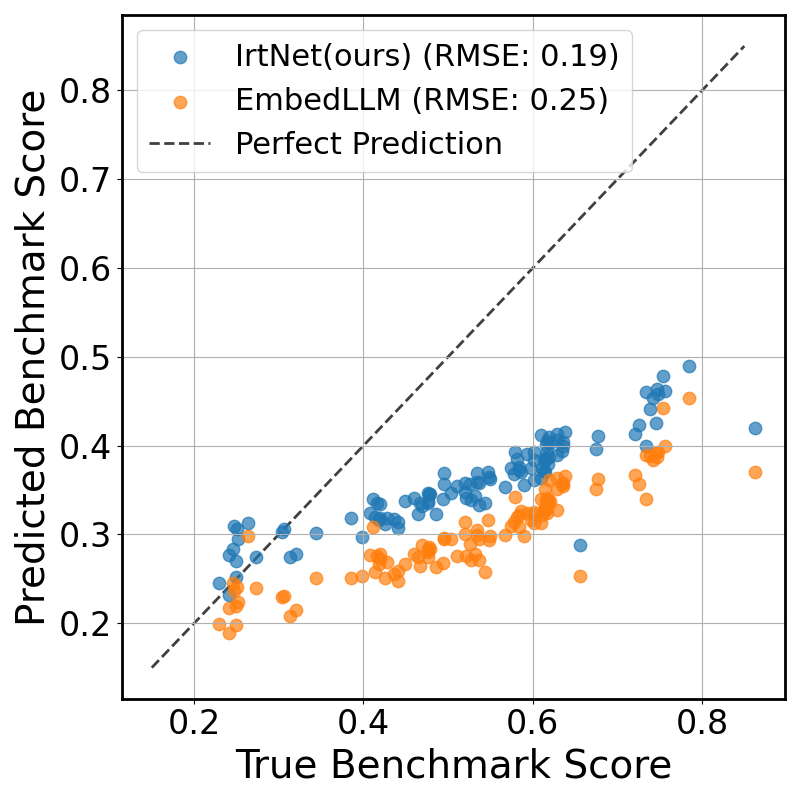}
        \caption{MMLU}
        \label{fig:sub1}
    \end{subfigure}
    \hfill 
    \begin{subfigure}[b]{0.32\columnwidth}
        \centering
        \includegraphics[width=\linewidth]{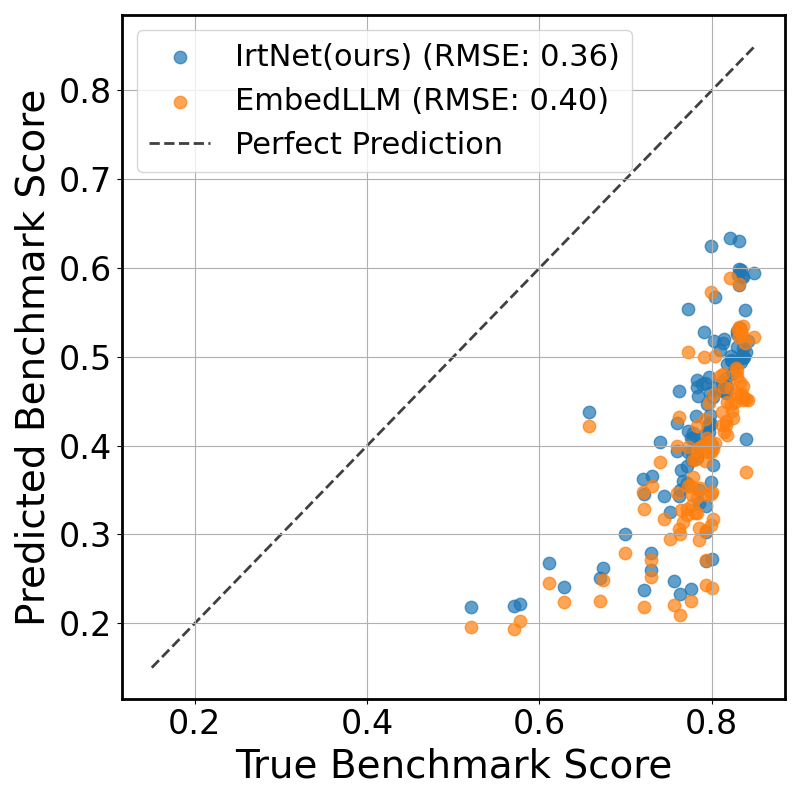}
        \caption{PIQA}
        \label{fig:sub2}
    \end{subfigure}
    \hfill 
    \begin{subfigure}[b]{0.32\columnwidth}
        \centering
    \includegraphics[width=\linewidth]{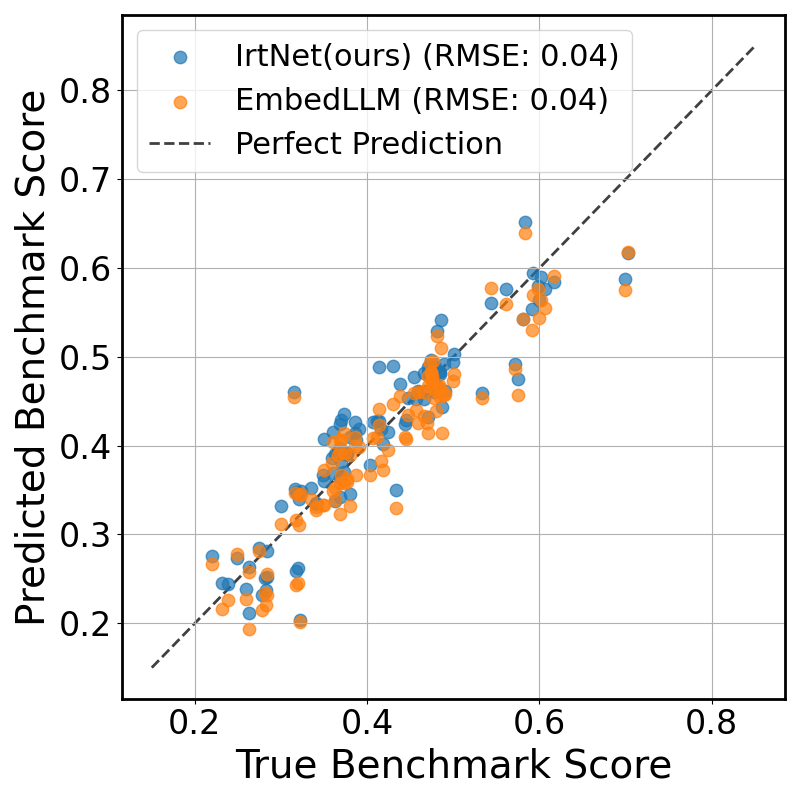}
        \caption{MedMCQA}
        \label{fig:sub3}
    \end{subfigure}
    \caption{Predicted vs. true benchmark scores (in [0-1]) on three OOD benchmarks.
    The scatter plots represent the predicted LLM scores by IrtNet and EmbedLLM. IrtNet's predictions (blue dots) align more closely with the perfect prediction diagonal line on MMLU and PIQA, which means lower prediction errors. IrtNet and EmbedLLM are tied on  MedMCQA with the predicted scores almost coinciding with the true scores.}
    \label{fig:three_images}
    \vspace{-5pt}
\end{figure}

Figure~\ref{fig:three_images} presents the specific prediction distributions on three representative benchmarks (MMLU, PIQA, and MedMCQA), vividly illustrating how IrtNet's predictions achieve a smaller prediction error compared to EmbedLLM or cluster very tightly around the true scores. 

\subsection{Ablation Study}
In this section, we investigate the role of the MoE layer in understanding queries. We ablate the MoE layer, replacing it with an MLP network of the same number of parameters, while keeping all other network structures unchanged. 
\begin{table}[!h]
    \centering
    \caption{Ablation study on the MoE layer of IrtNet. \textbf{Routing} denotes the model routing task and \textbf{Correctness} denotes the correctness prediction task.}
    \begin{tabular}{lll}
    \toprule
        \bf{Method}           & \bf{Routing} &  \bf{Correctness}\\
        \hline
        IrtNet            & $67.4$         & $72.2$   \\
        \quad w/o MoE     & $64.0_{\textcolor[HTML]{32B897}{\downarrow3.4}}$   & $71.3_{\textcolor[HTML]{32B897}{\downarrow0.9}}$        \\
    \bottomrule
    \end{tabular}
    \label{tab:ablation}
    \vspace{-10pt}
\end{table}

As shown in Table~\ref{tab:ablation}, after removing the MoE layer, IrtNet exhibits a significant performance drop in the model routing task and a slight decline in the correctness prediction task. The performance degradation in both tasks obviously suggests that the MoE layer is instrumental in learning more effective $\alpha_q$ and $\beta_q$ representations, thereby enhancing the network's overall prediction accuracy. Furthermore, the more substantial decline in the model routing task indicates that the MoE layer is particularly effective at understanding the relative ranking among models. This is likely because the MoE layer produces a more discriminative $\alpha_q$ representation, which more accurately captures the distinctions among model abilities.

\section{Interpretability Analysis}
\subsection{Understanding discrimination}
In our framework, the discrimination vector $\alpha_q$ provides a rich, quantitative profile of a query's intrinsic characteristic. A high value in a particular dimension of $\alpha_q$ signifies that the query places a strong demand on the corresponding latent ability, making a model's proficiency in that dimension more critical for a correct answer. Thus, the discrimination vector reflects which abilities a query is designed to test and how strongly it tests them. To validate this interpretation, we conduct a visualization experiment on the test set. We use the trained IrtNet to compute the discrimination vectors for  queries on the test set and then project these high-dimensional vectors into a two-dimensional space using t-SNE. 
\begin{figure}[tbh!]
    \centering
    \includegraphics[width=0.8\columnwidth]{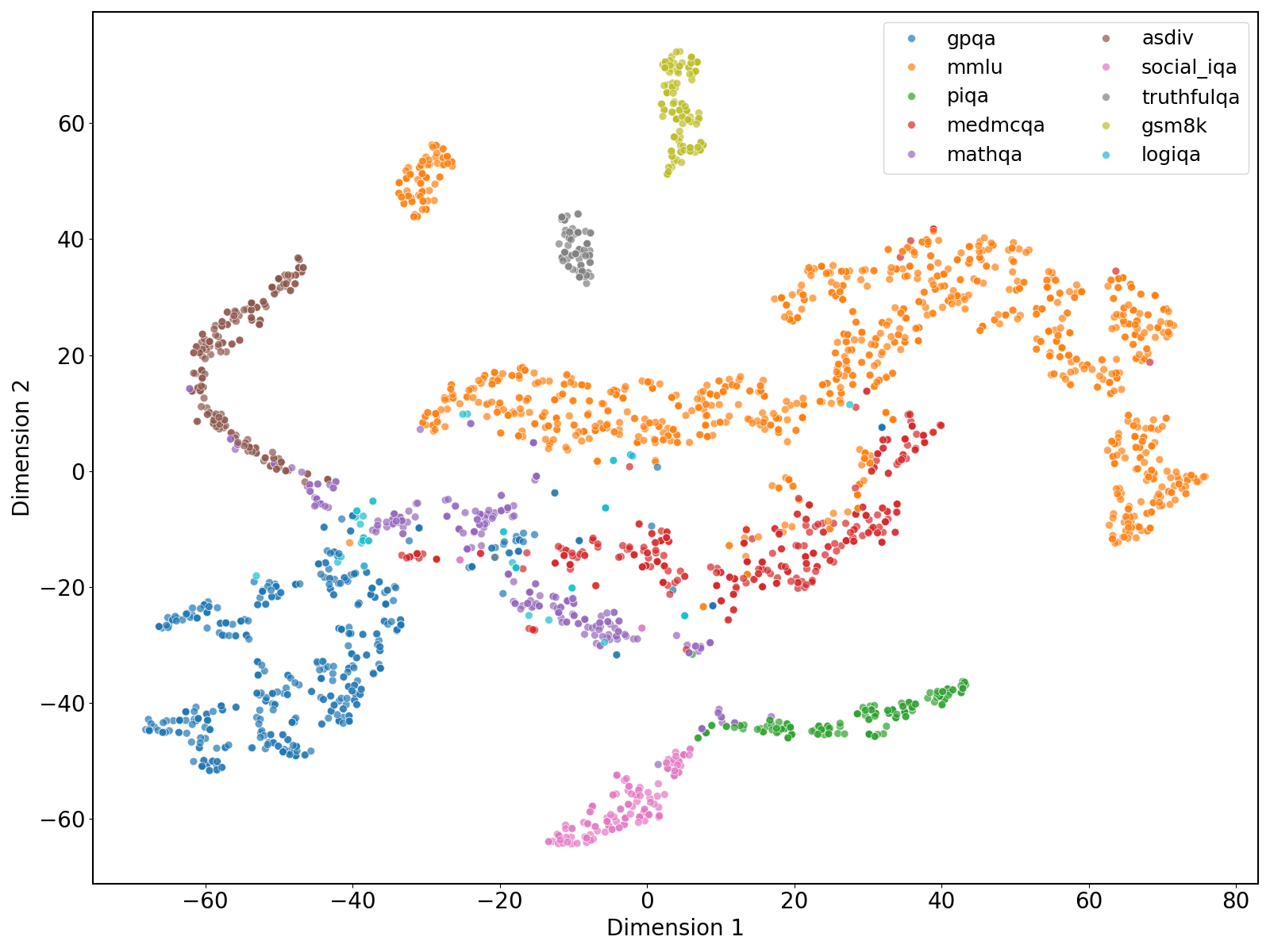}
    \caption{T-SNE visualization of learned query discrimination vectors $\alpha_q$.}
    \label{fig:discrimination}
    \vspace{-10pt}
\end{figure} 

Remarkably, despite IrtNet never being exposed to any dataset labels during training, the resulting visualization in Figure~\ref{fig:discrimination} reveals a clear and well-defined clustering structure. We observe that queries originating from the same benchmark naturally form distinct semantic groups in this learned space. This provides strong evidence that the learned discrimination vector $\alpha_q$ has successfully captured the unique demands of different query types. The spontaneous emergence of these clusters demonstrates that IrtNet has effectively modeled the distinct discriminative properties inherent to each semantic query group, mapping them into an interpretable space.

\subsection{Validating Difficulty}
We perform a quantitative analysis to validate that our learned difficulty parameter $\beta_q$ is a meaningful measure of a query's intrinsic challenge. For our experimental setup, we first establish an objective difficulty standard for each benchmark. This standard is defined as the average ground-truth accuracy achieved by the entire pool of 112 models on that dataset. A lower average score naturally corresponds to a higher objective difficulty. We then compare this objective standard against the average learned difficulty parameter $\beta_q$ for each respective benchmark. 

\begin{table}[!h]
    \centering
    \vspace{-5pt}
    \caption{Comparison of average model accuracy (in [0, 1]) and learned difficulty $\beta_q$ across benchmarks.}
    \small
    \begin{tabularx}{\textwidth}{l*{11}{>{\centering\arraybackslash}X}}
    \toprule
                & \bf{ASD} & \bf{GPQA} & \bf{Logi} & \bf{Math} & \bf{Social} & \bf{Truth} & \bf{Med} & \bf{GSM} & \bf{Mmlu} & \bf{PIQA} \\
        \midrule
        Average Accuracy & 0.04 & 0.21 & 0.29 & 0.33 & 0.34 & 0.36 & 0.42 & 0.42 & 0.53 & 0.78  \\
        Average $\beta_q$ & 1.69 & 0.71 & 0.45 & 0.40 & 0.33 & 0.32 & 0.16 & 0.17 & -0.08 & -0.78 \\
    \bottomrule
    \end{tabularx}
    \label{tab:difficulty}
\end{table}

As shown in Table~\ref{tab:difficulty}, our findings reveal an exceptionally strong negative correlation between the two: as the average model accuracy on a benchmark decreases (i.e., it gets harder), the learned $\beta_q$ value consistently increases. Quantitatively, the Pearson correlation coefficient between the objective average accuracy and the learned average difficulty $\beta_q$ is -0.9721. This near-perfect correlation provides compelling evidence that the difficulty parameter learned by IrtNet is not an arbitrary value, but a valid and reliable metric that accurately captures the empirical hardness of the queries.

\subsection{Probing LLM Embedding}
To investigate whether the learned LLM embedding $\theta_m$ captures meaningful model characteristics, we conduct a similarity validation experiment. Our hypothesis is that if the vectors are coherent, models sharing fundamental traits should be geometrically closer in the learned space. To test this, we partition our pool of models into distinct groups based on two criteria: model family, such as the Qwen and Llama families, and domain specialization for tasks like medicine, code, and math. We then calculate and compare two metrics: the average intra-community L2 distance, which measures the distance between models within a single group, and the average inter-community L2 distance, which measures the distance from that group's models to all outside models.

\begin{figure}[tbh!]
    \centering
    \vspace{-5pt}
    \includegraphics[width=\columnwidth]{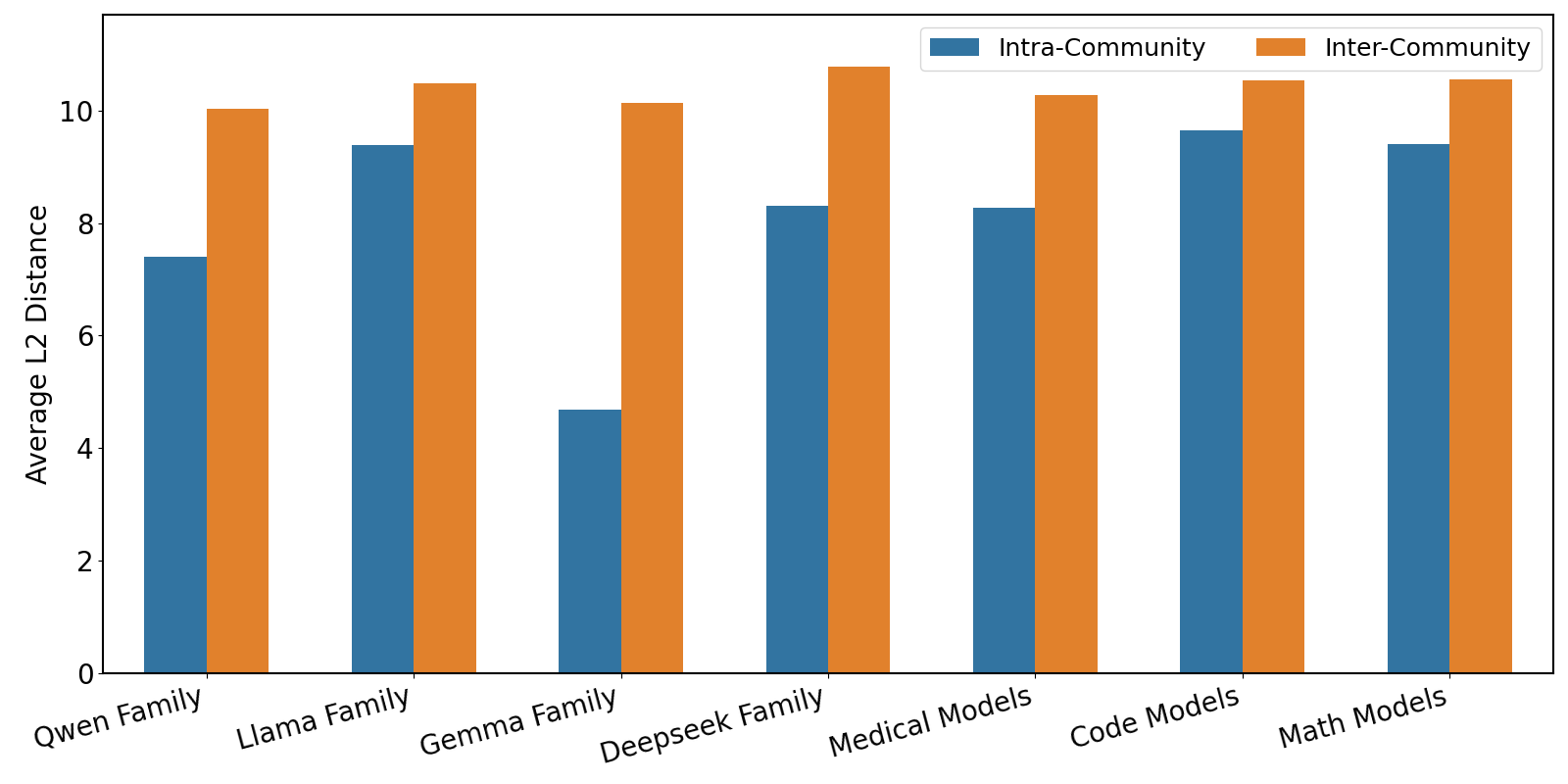}
    \caption{Comparison of intra-community and inter-community L2 distances for LLM embeddings.}
    \label{fig:distance}
\end{figure} 
Figure~\ref{fig:distance} provides compelling evidence for our hypothesis. Across all defined groups, the average intra-community distance is consistently and significantly smaller than the average inter-community distance. This clear geometric clustering holds true for models grouped by both architectural lineage, such as the Llama and Gemma families, or by functional specialization, like the groups of Code and Math models. This finding demonstrates that model embedding $\theta_m$ is a meaningful, well-structured representation that effectively encodes a model's specialized abilities.




\section{Related Work}
\paragraph{Representation Learning}
The concept of learning compact vector representations for complex, high-dimensional objects is a well-established paradigm in machine learning~\cite{}. For instance, models like SentenceTransformer~\citep{reimers2019sentence} and Qwen3-Embedding~\citep{zhang2025qwen3} effectively embed sentences into a dense embedding space;
in knowledge graphs, methods like TransE~\citep{bordes2013translating} learn low-dimensional embeddings for entities and relations.
More recently, EmbedLLM~\citep{embedllm} generalizes this paradigm to model LLM abilities with an encoder-decoder approach, demonstrating that learning a compact representation of LLM abilities can facilitate multiple downstream tasks. Our work is most closely related to EmbedLLM, but bears key conceptual and methodological differences. 
We apply IRT to model LLM abilities, providing a theory-driven approach to this paradigm, and introduce an MoE-based method to explicitly learn query characteristics (discrimination and difficulty) that EmbedLLM does not capture. Furthermore, our method leads to significant outperformance compared to EmbedLLM.

\paragraph{Model Routing}
Model routing for LLMs~\citep{shnitzer2023large,zooter,srivatsa2024harnessing,chen2504we} has emerged as a critical strategy for efficiently managing a diverse suite of models. Recent research in this field generally falls into two streams. The first emphasizes striking a balance between performance and computational efficiency~\citep{jiang2023llm,ong2024routellm,feng2024graphrouter,wang2025mixllm,zhang2025beyond}. The second prioritizes pushing model performance to the highest possible levels~\citep{zooter,routerDC,modelSAT,zhang2025avengers}. RouterDC~\citep{routerDC} proposes
a dual-contrastive learning framework to better align
queries and model representations. Model-SAT~\citep{modelSAT} creates a capability representation for each model through a lightweight aptitude test to select the most suitable model. The Avengers~\citep{zhang2025avengers} presents a training-free, clustering-based routing framework that selects the optimal model.
While our method does not specifically target model routing, the router developed using our approach significantly outperforms state-of-the-art routing methods. This demonstrates that accurately learning compact representations of model abilities, particularly when guided by IRT, play an important role in model routing.

\paragraph{Benchmark Prediction}
Evaluating LLMs on large benchmarks requires substantial computational and financial cost, which has motivated research on benchmark prediction that seeks to estimate overall performance from only a subset of representative data. Core-set selection methods such as Anchor Points \citep{vivek2023anchor} attempt to identify a small number of informative queries that preserve model rankings nearly as well as full benchmarks. Other work has shown that even simple random subsets combined with regression models can provide surprisingly strong estimates of average accuracy \citep{zhang2025benchmark}. More recently, IRT-based methods such as tinyBenchmarks \citep{polo2024tinybenchmarks} have demonstrated that psychometric modeling can reduce evaluation costs even further. 
While our approach to learning representations of model abilities can be applied to benchmark prediction, it differs from the above studies that focus on data selection strategies. In this work, we isolate the effect of our method by focusing on random data selection, but combining our approach with those selection strategies would be a promising direction for future work.





\section{Conclusion}
In this paper, we pioneer the application of item response theory (IRT) to formally model LLM abilities. We introduce IrtNet, a IRT-based framework that learns compact representations of LLM abilities. Based on a Mixture-of-Experts architecture, it jointly learns model embeddings alongside query difficulty and discrimination. Extensive Experiments demonstrate that IrtNet sets a new state-of-the-art in model routing and achieves highly data-efficient, more accurate benchmark prediction. Furthermore, the learned representations are also interpretable. The difficulty parameter strongly correlates with empirical results, and that model embeddings naturally cluster by family and function. Overall, IrtNet provides a robust and insightful tool for effective model evaluation, selection, and analysis in the growing LLM ecosystem.



\bibliography{iclr2026_conference}
\bibliographystyle{iclr2026_conference}

\end{document}